\documentclass{article} 

\usepackage{iclr2024_conference,times}
\usepackage{latexsym}
\usepackage{tabularx}
\usepackage{adjustbox}
\usepackage{multirow}
\usepackage{booktabs}
\usepackage{subfigure}
\usepackage{hyperref}
\usepackage{url}
\usepackage{float}
\usepackage{wrapfig}

\usepackage{soul}
\usepackage{changepage,threeparttable} 

\usepackage[T1]{fontenc}

\usepackage[utf8]{inputenc}

\usepackage{microtype}

\usepackage{inconsolata}

\usepackage{amsmath}

\title{What Happens When Small Is Made Smaller? Exploring the Impact of Compression on Small Data Pretrained Language Models}

\author{\vspace{2mm}\hspace{1cm}Busayo Awobade\thanks{These authors contributed equally to this work.},\hspace{1cm}  Mardiyyah Oduwole\footnotemark[1],\hspace{1cm} Steven Kolawole\footnotemark[1]\\
\hspace{5.5cm} \vspace{1mm}ML Collective \\
\texttt{\hspace{1cm}\{busayo.awobade, mardiyyah.oduwole, steven\}@mlcollective.org} \\
}

\begin{document}
\maketitle
\begin{abstract}
Compression techniques have been crucial in advancing machine learning by enabling efficient training and deployment of large-scale language models. However, these techniques have received limited attention in the context of low-resource language models, which are trained on even smaller amounts of data and under computational constraints, a scenario known as the "low-resource double-bind."
This paper investigates the effectiveness of pruning, knowledge distillation, and quantization on an exclusively low-resourced, small-data language model, AfriBERTa.  Through a battery of experiments, we assess the effects of compression on performance across several metrics beyond accuracy.
Our study provides evidence that compression techniques significantly improve the efficiency and effectiveness of small-data language models, confirming that the prevailing beliefs regarding the effects of compression on large, heavily parameterized models hold true for less-parameterized, small-data models.
\end{abstract}

\section{Introduction}
One of the most challenging aspects of working with large language models (LLMs) is their computational complexity \citep{zhang_when_2021}. With 340M parameters, even the BERT-large model is impractical for deployment on low-end devices with inadequate computational power \citep{treviso2022efficient}. Several architectural changes to make the BERT model more efficient have been made \citep{Jiao2019TinyBERTDB, Sanh2019DistilBERTAD, lan_albert_2023}. However, to achieve adequate performance on downstream tasks, LLMs require huge training corpora (billions of tokens), which are unavailable for most African languages \citep{nekoto2020participatory}. The omission of African languages from the pre-training phase of LLMs results in low performance in these languages, making NLP tasks participatory difficult \citep{kreutzer_quality_2022}. \citet{ahia2021low} termed this situation the "low-resource double-bind" to describe the coexistence of data and computation limitations on resources. This is a popular NLP setting for low-resource languages, although the performance trade-offs are understudied.

One of the most promising attempts to mitigate the sparse presence of low-resource African languages in model training is the creation of AfriBERTa, the first multilingual language model trained purely and from scratch on African languages with $<1$GB of data. AfriBERTa \citep{ogueji-etal-2021-small} beats competitive models like mBERT \citep{devlin_bert_2019} and XLM-R \citep{conneau2020unsupervised} on text categorization and NER tasks. Instead of depending on high-resource languages for transfer learning, AfriBERTa takes advantage of linguistic similarities between languages from low-resource environments to yield promising results, which is critical in determining the sustainability of language models trained on small datasets. However, with 126M parameters, the AfriBERTa-large model is still impractical for deployment on low-end devices with inadequate computational power. Moreover, little is known about the ability of a small-data, low-resource-focused model like AfriBERTa to generalize to unseen-before languages, considering its small size, and its ability to get "smaller" for efficient usage by resource-constrained users.

This research attempts to bridge the gap between a low-resource, small-data, high-performance multilingual language model and an ultra-efficient, deployable model for double-bind users. 
Our experimental results demonstrate that pruning achieves $\approx60\%$ reduction in model size with a minimal performance drop. Furthermore, generalization tests reveal varied outcomes, with some languages surpassing dense models even with extreme pruning. Distillation achieves compression rates between 22\% and 33\% with comparable performances.
Additionally, quantization reduces the model size by 64.08\%, inference time by 52.3\%, and even outperforms the baseline model in the F1 score for certain languages. Our contributions address the following questions:
\begin{enumerate}
    \item How tiny can we construct a small-data model using the knowledge distillation framework?
    \item What are the efficiency and generalization limits of pruning on a small-data model?
    \item What are the optimal reductions we can achieve in size and latency utilizing quantization?
    
\end{enumerate}


\paragraph{Related work}
First introduced by \cite{Hinton2015DistillingTK}, \citet{Sanh2019DistilBERTAD} demonstrated similar performances on downstream tasks with smaller language models pre-trained using distillation, which is faster at inference and suitable for edge devices. \citet{Jiao2019TinyBERTDB} proposed a transformer-specific distillation method, employing a two-stage learning framework with general and task-specific distillation using BERT.
\citet{Han2015DeepCC} reintroduced modern pruning as "network pruning." and the spotlight intensified with \citet{franklelottery} suggesting the existence of subnetworks within a dense neural network that matches or surpass the dense model's performance—termed "winning tickets." \citet{yu2019playing} and \citet{renda2020comparing} also found winning tickets early in training for Transformers and LSTMs.
\citet{chen2020lottery} and \citet{prasanna2020bert} also explored trainable subnetworks in pre-trained BERT models, locating matching subnetworks at 40\% to 90\% sparsity across various applications.
\citet{li2020train} demonstrated that heavily compressing large models resulted in higher accuracy than lightly compressing small models.
\citet{bai2022towards} also introduced post-training quantization for language models, minimizing training time, memory, and data consumption, while \citet{wang2022deep} achieved 16$\times$ compression by quantizing transformer backbones to 4-bit and applying 50\% fine-grained structural sparsity.
Additionally, \citet{xiao2022smoothquant} enabled 8-bit weight, 8-bit activation quantization for large language models, addressing activation outliers, and \citet{dettmers2022llm} used LLM.int8() on transformers with 16 or 32-bit weights for immediate inference using vector-wise quantization and mixed-precision decomposition. As far as we know, our work is the first to explore these techniques' efficacy in a low-resource double-bind setting.

\section{Approach}
This section details our experimental settings for the compression techniques we used to evaluate efficiency in our small-data, pre-trained model of choice. Further details on our training setup can be found in Appendix \ref{sec:training_setup}, and all settings stay consistent for all our experiments.

\paragraph{Data}
We use the AfriBERTa corpus for distillation; it comprises 11 African languages. We use the MasakhaNER dataset \citep{Adelani2021MasakhaNERNE}, a Named Entity Recognition dataset that spans 10 African languages, for task-specific compression evaluation.

\paragraph{Model Training} 
This study uses the AfriBERTa models, \emph{Base} and \emph{Large}, which are based on the XLM-R architecture. For the different compression strategies we explore, we use either or both the large variant and base variant for our study.
The final results were averaged over three training runs with different training seeds for all reported results. We notice an insignificant standard deviation and distribution between the results of the different seeds.

\paragraph{Tasks Evaluation}
The evaluation of the compressed models in this study focuses on the NER task due to its significant relevance in downstream applications such as question answering and information extraction \citep{tjong2003introduction}. We adopt the F1 score as our primary evaluation metric, and the evaluation dataset is the MasakhaNER dataset \citep{Adelani2021MasakhaNERNE}. 

\subsection{Efficiency Evaluation} 
\paragraph{Distillation}
We apply both task-agnostic and task-specific distillation approaches on the base and large variants of the AfriBERTa model. We distil knowledge from the pre-trained AfriBERTa model into relatively smaller models for task-agnostic distillation and evaluate them for the NER downstream task. For the task-specific distillation, we distil knowledge from a model fine-tuned for the NER downstream task into smaller models.
\paragraph{Pruning}
Our pruning experiments at various sparsity levels range from $10\%$ to $95\%$. We prune the model before, after, and during fine-tuning to examine the impact of pruning at each phase of the training process. We also examine the computational efficiency of the pruned models by measuring their inference time on the test data at all sparsity levels.
Furthermore, we evaluate the generalization capabilities of the pruned models by fine-tuning and testing them on out-of-distribution (OOD) data. We compare the performance of the pruned models to the original dense model to assess any influence of pruning on cross-lingual knowledge transfer and the model's level of generalizability, using MasakhaNER 2.0 \citep{adelani2022masakhaner} and MSRA NER \citep{feng2006chinese} dataset.
\paragraph{Quantization} 
We examine the effects of quantization on the large AfriBERTa model, which has been fine-tuned for Named Entity Recognition (NER) tasks. Our study utilizes two quantization approaches. The first is the \textbf{LLM.int8()} method, which uniformly converts the model's weights, activations, and attention mechanisms to an 8-bit integer (int8) format. The second approach is \textbf{dynamic quantization}, which dynamically converts the weights of linear layers from floating-point to integer data types at runtime and quantizes the activation layers during CPU-based inference.

\begin{table*}[t]
\centering
\caption{\textbf{Average results for the distilled models on NER task across 10 languages}. The best student variant for each teacher is highlighted, and the best variant for each strategy is underlined.}
\label{tab:distilation_results}
\scriptsize
\begin{adjustbox}{width=\textwidth}
\begin{tabular}{llrrlrrrrrrrrrrr}
\toprule
\textbf{Distillation strategy} &         \textbf{Teacher} &  \textbf{\#Layers} &  \textbf{\#Att. Heads} & \textbf{\#Params} &     \textbf{amh} &     \textbf{hau} &     \textbf{ibo} &     \textbf{kin} &     \textbf{lug} &     \textbf{luo} &     \textbf{pcm} &     \textbf{swa} &     \textbf{wol} &     \textbf{yor} &      \textbf{avg} \\
\midrule

        Task-agnostic &  AfriBERTa-base &        4 &            4 &     83M & 64.96 & 87.28 & 83.58 & 68.15 & 74.57 & 63.02 & 78.92 & 83.89 & 55.46 & 73.62 & 73.35 \\
                      &                 &        4 &            6 &     83M & 64.23 & 87.34 & 83.84 & 67.59 & 74.60 & 60.00 & 79.40 & 84.00 & 57.21 & 73.38 & 73.16 \\
                      &                 &        6 &            4 &     97M & 65.96 & 87.60 & 85.55 & 70.16 & 75.90 & 64.61 & 81.65 & 85.48 & 57.70 & 74.82 & 74.94 \\
                      &                 &        6 &            6 &     97M & 66.92 & 87.91 & 85.28 & 69.81 & 77.19 & 68.40 & 81.72 & 85.08 & 60.28 & 75.47 & \textbf{75.81} \\
                     \cmidrule(l){2-16}
                      & AfriBERTa-large &        4 &            4 &     83M &  65.28 &  87.28 &  84.15 &  68.83 &  73.82 &  63.79 &  79.80 &  84.13 &  56.30 &  73.43 &  73.68 \\
                      &                 &        4 &            6 &     83M & 65.25 & 87.62 & 84.28 & 68.82 & 74.66 & 62.60 & 78.88 & 84.07 & 55.11 & 73.85 & 73.51 \\
                      &                 &        6 &            4 &     97M & 69.38 & 88.25 & 85.08 & 69.49 & 75.44 & 63.87 & 82.71 & 85.87 & 56.42 & 73.89 & 75.04 \\
                      &                 &        6 &            6 &     97M &  71.98 &  88.72 &  85.76 &  71.76 &  78.30 &  68.10 &  84.24 &  87.07 &  61.45 &  77.61 &  \underline{\textbf{77.50}} \\
       \midrule
        Task-specific &  AfriBERTa-base &        4 &            4 &     83M & 65.04 & 87.12 & 83.13 & 67.62 & 75.01 & 62.90 & 78.19 & 83.96 & 54.04 & 69.22 & 72.62 \\
                      &                 &        4 &            6 &     83M & 65.52 & 87.27 & 83.93 & 68.18 & 75.56 & 63.78 & 79.03 & 83.70 & 57.46 & 69.98 & 73.44 \\
                      &                 &        6 &            4 &     97M & 67.28 & 87.27 & 85.72 & 71.68 & 77.18 & 66.58 & 81.85 & 84.92 & 60.20 & 74.78 & \textbf{75.75} \\
                      &                 &        6 &            6 &     97M & 69.45 & 88.23 & 85.47 & 69.88 & 74.90 & 65.79 & 82.27 & 85.36 & 59.12 & 76.25 & 75.67 \\
                      \cmidrule(l){2-16}
                      & AfriBERTa-large &        4 &            4 &     83M & 65.66 & 87.60 & 83.42 & 67.38 & 74.28 & 62.37 & 79.62 & 83.78 & 55.72 & 72.89 & 73.27 \\
                      &                 &        4 &            6 &     83M & 71.20 & 88.27 & 84.66 & 70.70 & 77.11 & 65.58 & 82.09 & 86.06 & 58.00 & 76.21 & 75.99 \\
                      &                 &        6 &            4 &     97M & 69.38 & 88.25 & 85.08 & 69.49 & 75.44 & 63.87 & 82.71 & 85.87 & 56.42 & 73.89 & 75.04 \\
                      &                 &        6 &            6 &     97M & 72.58 & 88.33 & 86.05 & 71.16 & 78.56 & 69.87 & 84.03 & 86.32 & 61.49 & 76.66 & \underline{\textbf{77.51}} \\
\bottomrule
\end{tabular}
\end{adjustbox}
\end{table*}

\begin{table*}[!htp]\centering
\caption{\textbf{Comparison of NER results between the teachers and the best students.} The underlined scores are instances where the distilled model outperformed any of the teacher models.}\label{tab: teacher vs student}
\scriptsize
\begin{adjustbox}{width=\textwidth}
\begin{tabular}{lrrrrrrr}\toprule
\textbf{Language} &\textbf{AfriBERTa-base} &\textbf{AfriBERTA-large} &\textbf{Distiled AfriBERTa-base} &\textbf{Distiled AfriBERTa-base} &\textbf{Distiled AfriBERTa-large} &\textbf{Distiled AfriBERTa-large} \\
\textbf{} &\textbf{(Teacher)} &\textbf{(Teacher)} &\textbf{(Task agnostic)} &\textbf{(Task specific) } &\textbf{(Task agnostic)} &\textbf{(Task Specific)} \\
\textbf{} &\textbf{<111M} &\textbf{<126M>} &\textbf{<97M>} &\textbf{<97M>} &\textbf{<97M>} &\textbf{<97M>} \\\midrule
amh &71.8 &\textbf{73.28} &69.45 &66.92 &71.98 &\ul{72.58} \\
hau &90.01 &\textbf{90.17} &88.23 &87.91 &88.72 &88.33 \\
ibo &86.63 &\textbf{87.38} &85.47 &85.28 &85.76 &\ul{86.05} \\
kin &69.91 &\textbf{73.78} &69.88 &69.81 &\ul{71.76} &71.16 \\
lug &76.44 &\textbf{78.85} &74.9 &77.19 &78.3 &\ul{78.56} \\
luo &67.31 &\textbf{70.23} &65.79 &68.4 &68.1 &\ul{69.87} \\
pcm &82.92 &\textbf{85.7} &82.27 &81.72 &\ul{84.24} &84.03 \\
swa &85.68 &\textbf{87.96} &85.36 &85.08 &\ul{87.07} &86.32 \\
wol &60.1 &\textbf{61.81} &59.12 &60.28 &61.45 &\ul{61.49} \\
yor &76.08 &\textbf{81.32} &76.25 &75.47 &\ul{77.61} &76.66 \\
\midrule
\textbf{avg} &76.688 &\textbf{79.048} &75.672 &75.806 &77.499 &\ul{77.505} \\
\bottomrule
\end{tabular}
\end{adjustbox}
\end{table*}
\section{Results and Discussion}

\subsection{How small can we make these language models?}
Using our distillation methods, we achieve up to 31\% compression while maintaining competitive results, with only a 7\% performance drop for our least-performing model and only a 1.9\% decline compared to the best-performing AfriBERTa model at 22\% compression, as shown in table \ref{tab: teacher vs student}. We notice only marginal differences between the teachers' and students' performances in some languages. We also see the student model trained by the large teacher outperform the base teacher in specific languages.
Additionally, our task-agnostic models outperform the task-specific models in terms of F1 score, but with relatively minimal differences.

\subsection{Which is the best teacher: Base vs Large?}
Although there is a performance decline of roughly 1.9\% from the AfriBERTa-large baseline, we discover that the AfriBERTa-large model produced the student with the best grade. However, the best-performing student by the base model only showed a performance decline of 1.3\% between the original scores of the base model and the student scores (see Table \ref{tab: teacher vs student}). According to the results, the base model is comparatively better at imparting most of its knowledge to its students, even though the larger model creates the best overall student. Additionally, as the attention head and layer ratio reduce, the students being taught using the base model catch up to those taught by the larger model with no discernible difference in performance, as seen in Table \ref{tab:distilation_results}.
Our results suggest that the selected instructor model significantly influences the performance of student models.

\begin{figure*}[t]
\centering
\subfigure[Pruning before fine-tuning]{
    \includegraphics[width=0.45\textwidth]{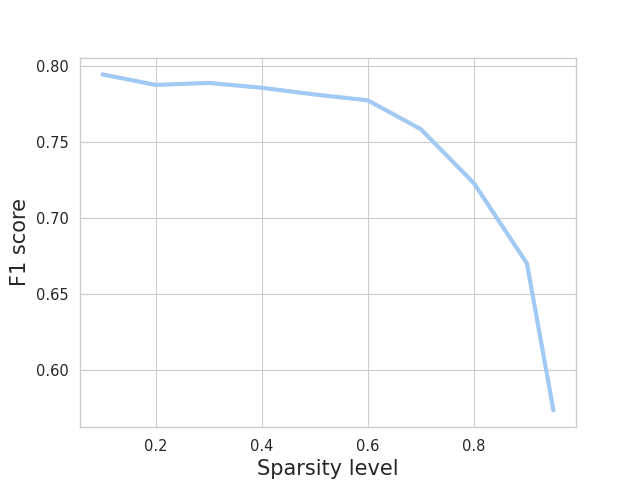}
}
\hfill
\subfigure[Pruning after fine-tuning]{
    \includegraphics[width=0.45\textwidth]{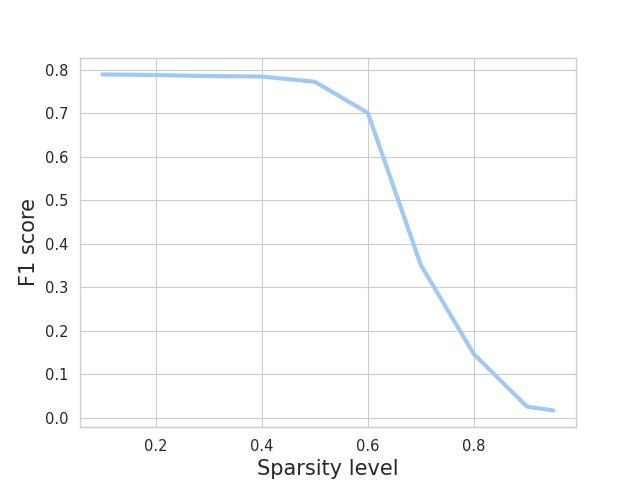}
}
\caption{Pruning \textbf{before vs after} fine-tuning: F1 score of sparsified models, averaged across languages.}

\label{fig:before_vs_after_finetuning}
\end{figure*}

\subsection{How does pruning before and after fine-tuning affect model performance?}
As seen in Figure \ref{fig:before_vs_after_finetuning}, our results show that pruning before fine-tuning produces fairly consistent performance with the dense model up to a sparsity level of 60\%. However, above this threshold, the model's performance gradually declines. Still, it remains competitive even at 80\% sparsity, with an average F1 score of over 70\% (detailed tables and charts in Appendix \ref{sec:appendixpruninga}). When fine-tuning is performed before pruning, however, our results demonstrate that the model's performance stays firmly on par with, and even exceeds, the performance of the dense model up to 50\% sparsity. However, when the sparsity level increases, especially from the 70\% sparsity, we find a dramatic deterioration in performance. It is worth mentioning that both methods have advantages and disadvantages. Pruning before fine-tuning results in more stable and predictable performance at greater sparsity levels, making it a feasible alternative for applications requiring high sparsity levels. On the other hand, pruning after fine-tuning can greatly improve the model's performance at lower sparsity levels, making it a better technique for applications that emphasize high accuracy.

\subsection{Exploring the limits of pruning for small-data pre-trained Models}
In our findings (see Appendix \ref{sec:appendixpruninga}), we notice that despite the limitations of AfriBERTa's training data and architecture, we find constant performance up to 50\% and 60\% sparsity. Notably, certain languages maintain a moderate degree of performance even at 95\% sparsity, suggesting that the model might have a certain level of robustness to pruning. Nonetheless, we notice dramatic reductions in performance for several languages, such as Yoruba and Luganda, at this level. This might be attributable to these languages' particular characteristics, such as their high inflectional complexity and the sparse nature of their datasets, which may make them more prone to pruning-induced degeneration.
To confirm the robustness of our results, we also explored cross-lingual transfer on AfriBERTa leveraging pruning, as detailed in Appendix \ref{sec:appendixpruningb}. While our results show that aggressive pruning is possible for small-data pre-trained models, it is also critical to take into account the unique qualities of each language and dataset when calculating the ideal sparsity level for pruning.

\subsection{How does Pruning affect Out-of-Domain Generalization?}
Our findings (Figure \ref{fig:ood_performance}) reveal that pruning can positively impact OOD generalization for some languages, while for others, the benefits are limited. Surprisingly, for many languages, including Swahili, the performance of the pruned models remains consistent with or surpasses that of the original dense model up to a sparsity level of 60\%, with $\approx90\%$ for Swahili.
However, for languages such as Yoruba, which exhibit a higher level of linguistic complexity, the performance is lower even for the dense model, with an F1 score of around 60\%, highlighting the challenge of compressing models with complex linguistic structures.

\subsection{Effectiveness of Quantization on Model Efficiency}
\begin{table}[t]
    \centering
    \begin{minipage}{0.4\textwidth} 
        \centering
        \small
        \caption{F1 Scores by Language and Quantization Methods.}
        \label{tab:f1_scores}
        \begin{tabular}{|c|c|c|c|}
            \hline
            Language & Baseline & Dynamic & LLM.int8() \\
            \hline
            Amh & \textbf{73.36} & 68.02 & 73.28 \\
            Hau & 89.93 & 85.35 & \textbf{89.95} \\
            Ibo & \textbf{86.96} & 82.21 & 86.88 \\
            Kin & \textbf{73.98} & 61.58 & 73.91 \\
            Lug & 79.78 & 68.94 & \textbf{79.83} \\
            Luo & \textbf{70.04} & 42.40 & 69.77 \\
            Pcm & \textbf{85.23} & 74.37 & 85.18 \\
            Swa & 87.89 & 84.58 & \textbf{87.93} \\
            Wol & \textbf{61.73} & 47.36 & 61.71 \\
            Yor & \textbf{80.76} & 65.10 & 80.74 \\
            \hline
        \end{tabular}
    \end{minipage}\hfill
    \begin{minipage}{0.5\textwidth} 
        \centering
        \small
        \caption{Inference Time Comparison (ms) for the quantization Methods and the baseline model.} 
        \label{tab:inference_time}
        \begin{tabular}{|c|c|c|c|}
            \hline
            Language & Baseline & Dynamic & LLM.int8() \\
            \hline
            amh & 26.01 & \textbf{12.78} & 13.27 \\
            hau & 31.08 & 19.99 & \textbf{13.31} \\
            ibo & 31.84 & 21.67 & \textbf{15.03} \\
            kin & 27.19 & 20.95 & \textbf{16.85} \\
            lug & 21.10 & 12.35 & \textbf{10.62} \\
            luo & 22.40 & 5.53 & \textbf{5.47} \\
            pcm & 41.70 & 17.96 & \textbf{16.34} \\
            swa & 35.50 & 20.14 & \textbf{17.37} \\
            wol & 25.38 & 20.95 & \textbf{14.78} \\
            yor & 34.45 & 23.14 & \textbf{18.36} \\
            \hline
        \end{tabular}
    \end{minipage}
\end{table}

Our results, as shown in Table \ref{tab:f1_scores}, show that the LLM.int8() quantization method generally outperformed the dynamic quantization method across all languages, with an average decrease in the F1-score of just 4.7\%. Moreover, our findings suggest that for some languages, such as Swahili, Luganda, and Hausa, LLM.int8() may be preferable to the original dense model. 

Model size reduced varyingly across languages, with dynamic quantization resulting in a 42.44\% reduction and LLM.int8() resulting in a 64.08\% reduction. There is no one-size-fits-all solution when it comes to quantization. The performance of quantized models depends on various factors, such as the language, the type of data being processed, and the adapted quantization technique.

Table \ref{tab:inference_time} shows that quantization can significantly reduce inference time for all languages. For example, in the case of Amharic, quantization lead to a 50\% reduction in inference time compared to the baseline model. Similarly, for Hausa and Swahili, quantization resulted in a 35\% and 40\% reduction in inference time, respectively. An average reduction of 40.9\% for dynamic quantization and 52.3\% for LLM.int8() was observed. These findings suggest that quantization effectively optimises small data-pre-trained models for deployment on devices with limited resources.

\section{Conclusion and Future Work}
This study investigates the effectiveness of pruning, knowledge distillation, and quantization on a small-data language model, AfriBERTa, trained on low-resource languages. Our findings indicate that compression techniques can significantly improve the efficiency and effectiveness of small-data language models. Also, we identify the importance of balancing the attention head and hidden layers when using knowledge distillation to compress small-data language models. Additionally, further experiments with different variations of quantization strategies yield results comparable to the original models. Our study balances compressed small-data language models' efficiency-accuracy tradeoff and generalization capabilities.

However, our work's novelty lies in applying existing compression techniques to a low-resource setting. We do not introduce new techniques or approaches but adapt and evaluate existing methods. Moreover, while NER is a crucial NLP task, a focus for future work is to explore the applicability of our findings to other NLP tasks.

\section*{Acknowledgements}
We thank the ML Collective community for the generous computational support, as well as helpful discussions, ideas, and feedback on experiments.

\bibliographystyle{plainnat}
\bibliography{main_bib}

\appendix

\section{Training Setup}
\label{sec:training_setup}

\subsection{Data}     
The AfriBERTa corpus is a multilingual dataset comprising 11 African languages. The data was primarily sourced from the BBC news\footnote{\url{https://www.bbc.co.uk/ws/languages} (scraped up to January 17, 2021)} and the Common Crawl Corpus \citep{conneau2020unsupervised}. Although relatively small at 0.91 GB, it was specifically engineered to present a first-of-its-kind attempt to train a multilingual language model exclusively on low-resource languages. Table \ref{tab:afriberta-corpus} shows language-specific information and token details.

The MasakhaNER dataset \citep{Adelani2021MasakhaNERNE} is used for task-specific compression and evaluation. It is a Named Entity ecognition dataset comprising PER, ORG, LOC, and DATE entities annotated for the 10 African languages. It was used to evaluate the original AfriBERTa model. The languages contained in the dataset are Amharic, Hausa, Igbo, Kinyarwanda, Luganda, Luo, Nigerian Pidgin, Swahili, Wolof, and Yorùbá. Table \ref{table:masakhaner} summarizes the dataset's details, including the languages included in the dataset.

\subsection{Data Preprocessing}

\begin{table}[t]
\centering
\caption{AfriBERTa corpus \citep{ogueji-etal-2021-small}: Size of each language in the
dataset covering numbers of sentences, tokens, and uncompressed disk size.}
\label{tab:afriberta-corpus}
\small
\begin{tabular}{lrrr}
\hline
\textbf{Language} & \textbf{\# Sent.} & \textbf{\# Tok.} & \textbf{Size (GB)} \\ \hline
Afaan Oromoo & 410,840 & 6,870,959 & 0.051 \\
Amharic & 525,024 & 1,303,086 & 0.213 \\
Gahuza & 131,952 & 3,669,538 & 0.026 \\
Hausa & 1,282,996 & 27,889,299 & 0.150 \\
Igbo & 337,081 & 6,853,500 & 0.042 \\
Nigerian Pidgin & 161,842 & 8,709,498 & 0.048 \\
Somali & 995,043 & 27,332,348 & 0.170 \\
Swahili & 1,442,911 & 30,053,834 & 0.185 \\
Tigrinya & 12,075 & 280,397 & 0.027 \\
Yorùbá & 149,147 & 4,385,797 & 0.027 \\ 
\textbf{Total} & \textbf{5,448,911} & \textbf{108,800,600} & \textbf{0.939} \\ \hline
\end{tabular}
\end{table}

Similar to the original preprocessing step for the AfriBERTa model \citep{ogueji-etal-2021-small}, we use the AfriBERTa tokenizer. We also remove lines that are empty or contain only punctuations to ensure that the dataset is clean and contains meaningful text. We also enforce a minimum length restriction by retaining sentences with more than 11 tokens. This step helps to filter out concise sentences that may not provide enough context for the model to learn effectively. 
Furthermore, we take preprocessing steps significant to the performance of the AfriBERTa model and DistilBERT \citep{Sanh2019DistilBERTAD}. We use the entire pre-training corpus since it's already a small amount of data of 1GB in size. 

\begin{table}[t]
    \centering
    \caption{Model architecture details for AfriBERTa-base and AfriBERTa-large.}
    \label{tab:afriberta-models}
    \small
    \begin{tabular}{lccc}
        \hline
        \textbf{Model} & \textbf{\#Params} & \textbf{\#Layers} & \textbf{\#Att. Heads} \\ \hline
        AfriBERTa-base & 111M & 8 & 6 \\
        AfriBERTa-large & 126M & 10 & 6 \\ \hline
    \end{tabular}
\end{table}

\subsection{NER}
Adapted from the \href{https://github.com/castorini/afriberta/blob/main/run_all.sh}{AfriBERTa} experiment setup 

\textbf{learning rate}: 5e-5

\textbf{max sequence length}: 164

\textbf{batch size}: 16

\textbf{num` epochs}:50

\textbf{Tokenizer}: `afriberta original tokenizer` \citep{ogueji-etal-2021-small}

\textbf{optimizer}: Adam \citep{Kingma2014AdamAM} 

\textbf{training seeds}: [1, 3, 5]

\section{Distillation}
\label{sec:distillation}
 We experiment using task-agnostic and task-specific distillation approaches on both the base and significant variants of the AfriBERTa model. 

\paragraph{Task-Agnostic Distillation}

Task-agnostic distillation involves distilling a large pre-trained language model into a smaller model that is not optimized for any specific downstream task. Knowledge from the teacher model was used to pretrain the student model which is then further fine-tuned on a downstream task, in our case here, the NER downstream task.

\paragraph{Task-Specific Distillation}
The already fine-tuned teacher model was used to teach the already distilled student model on a downstream task.
Task-specific distillation involves fine-tuning the pre-trained model on the target task and distilling knowledge from the fine-tuned model into the already distilled pretrained student model.

\subsubsection{Task agnostic}
Adapted from \href{https://github.com/huggingface/transformers/tree/main/examples/research_projects/distillation}{HuggingFace distillation}

\textbf{temperature}: [2, 3, 6]
\subsubsection{Task specific}
Adapted from \href{https://github.com/airaria/TextBrewer/tree/master/examples/conll2003_example}{TextBrewer}.   

\textbf{temperature}: 8

\begin{table*}[t]
\centering
\caption{Statistics of MasakhaNER datasets \cite{Adelani2021MasakhaNERNE} including their source, number of sentences in each split, number of annotators, and number of entities of each label type, combined with information on language, family, number of speakers \citep{eberhard2019ethnologue}, and African regions. Adapted from \citep{Adelani2021MasakhaNERNE}}
\label{table:masakhaner}
\begin{adjustbox}{width=\textwidth}
\begin{tabular}{lcccccccccc}
\hline
\textbf{Language} & \textbf{Family} & \textbf{Speakers} & \textbf{Region} & \textbf{Data Source} & \textbf{Train/Dev/Test} & \multirow{2}{*}{\shortstack{\textbf{\#}\\\textbf{Anno}}} & \textbf{PER} & \textbf{ORG} & \textbf{LOC} & \textbf{DATE} \\
 & & & & & & & & & & \\
\hline 
Amharic & Afro-Asiatic-Ethio-Semitic & 33M & East & DW \& BBC & 1750/250/500 & 4 & 730 & 403 & 1,420 & 580 \\
Hausa & Afro-Asiatic-Chadic & 63M & West & VOA Hausa & 1903/272/545 & 3 & 1,490 & 766 & 2,779 & 922 \\
Igbo & Niger-Congo-Volta-Niger & 27M & West & BBC Igbo & 2233/319/638 & 6 & 1,603 & 1,292 & 1,677 & 690 \\
Kinyarwanda & Niger-Congo-Bantu & 12M & East & IGIHE news & 2110/301/604 & 2 & 1,366 & 1,038 & 2,096 & 792 \\
Luganda & Niger-Congo-Bantu & 7M & East & BUKEDDE news & 2003/200/401 & 3 & 1,868 & 838 & 943 & 574 \\
Luo & Nilo Saharan & 4M & East & Ramogi FM news & 644/92/185 & 2 & 557 & 286 & 666 & 343 \\
Nigerian-Pidgin English & Creole & 75M & West & BBC Pidgin & 2100/300/600 & 5 & 2,602 & 1,042 & 1,317 & 1,242 \\
Swahili & Niger-Congo-Bantu & 98M & Central \& East & VOA Swahili & 2104/300/602 & 6 & 1,702 & 960 & 2,842 & 940 \\
Wolof & Niger-Congo-Senegambia & 5M & West \& NW & Lu Defu Waxu \& Saabal & 1871/267/536 & 2 & 731 & 245 & 836 & 206 \\
Yorùbá & Niger-Congo-Volta-Niger & 42M & West & GV \& VON news & 2124/303/608 & 5 & 1,039 & 835 & 1,627 & 853 \\
\hline
\end{tabular}
\end{adjustbox}
\end{table*}

\section{Additional Results References for Distillation}
\begin{figure*}[t!]
\centering
\caption{Performance (F1 score) during task agnostic and task-specific distillation across different languages in the NER task.}
\subfigure[Task agnostic]{
    \includegraphics[width=0.45\linewidth]{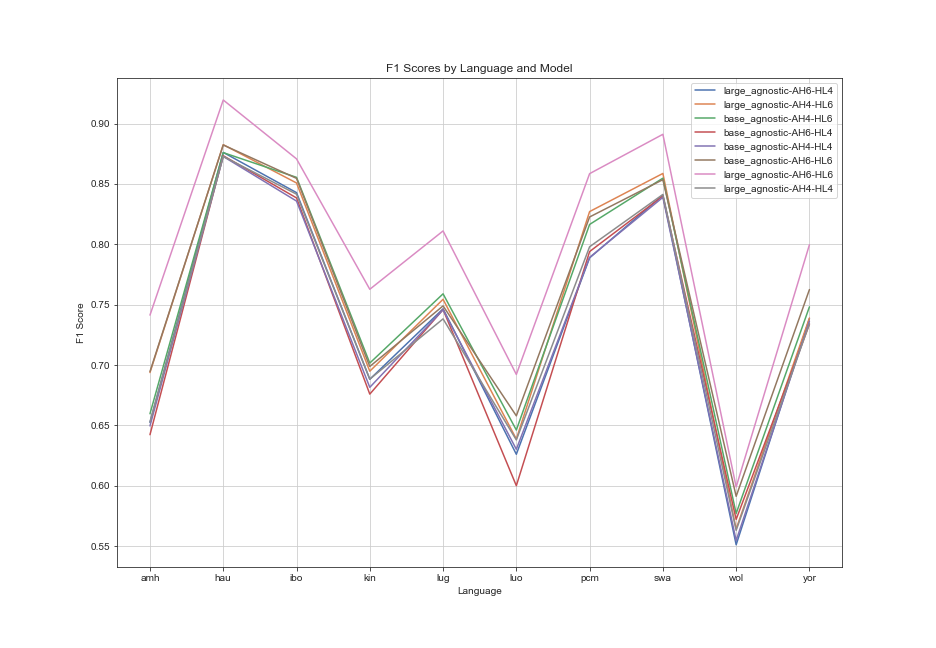}
}
\hfill
\subfigure[Task specific]{
    \includegraphics[width=0.45\linewidth]{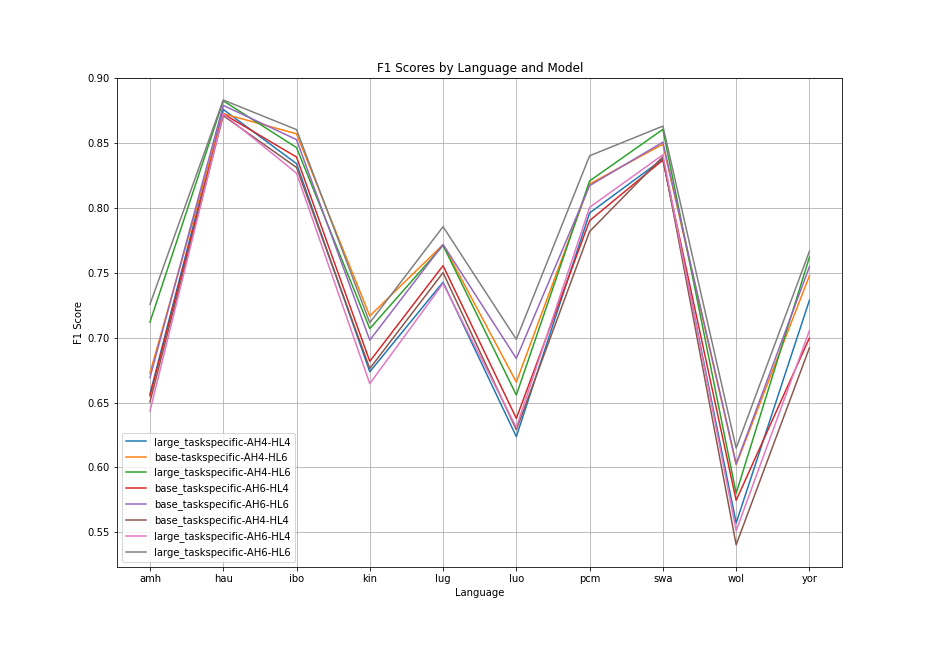}
}
\label{fig:agnostic_v_specific}
\end{figure*}

\begin{figure*}[t!]
\centering
\subfigure[Base teacher vs Students]{
    \includegraphics[width=0.45\linewidth]{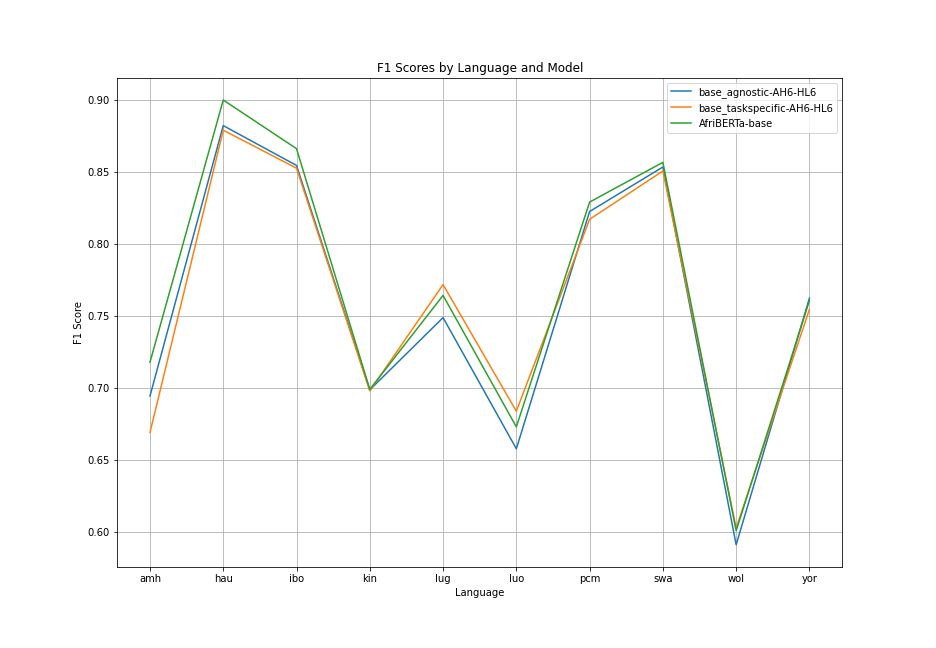}
}
\hfill
\subfigure[Large teacher vs Students]{
    \includegraphics[width=0.45\linewidth]{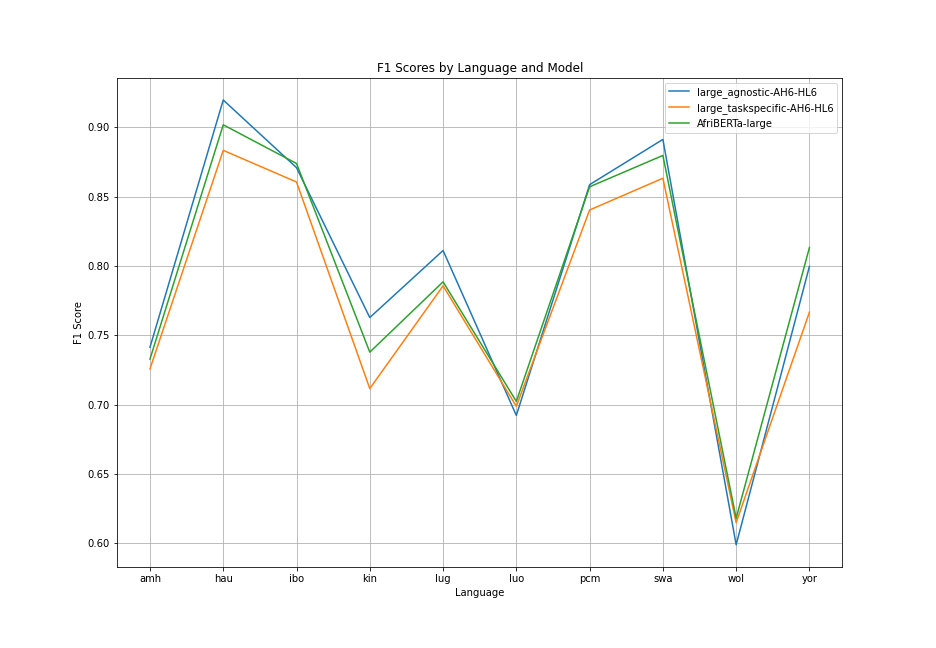}
}
\caption{Performance comparison between students and teachers (distillation).}
\label{fig:teacher_v_student}
\end{figure*}

See table \ref{tab:distilation_results} and figures \ref{fig:agnostic_v_specific} \& \ref{fig:teacher_v_student} for references.

\label{sec:appendixdistillation}

\begin{figure*}[t]
    \caption{A chart of the F1 scores for the best-performing student models and the teacher models across each language in the NER task.}
    \includegraphics[width=\linewidth]{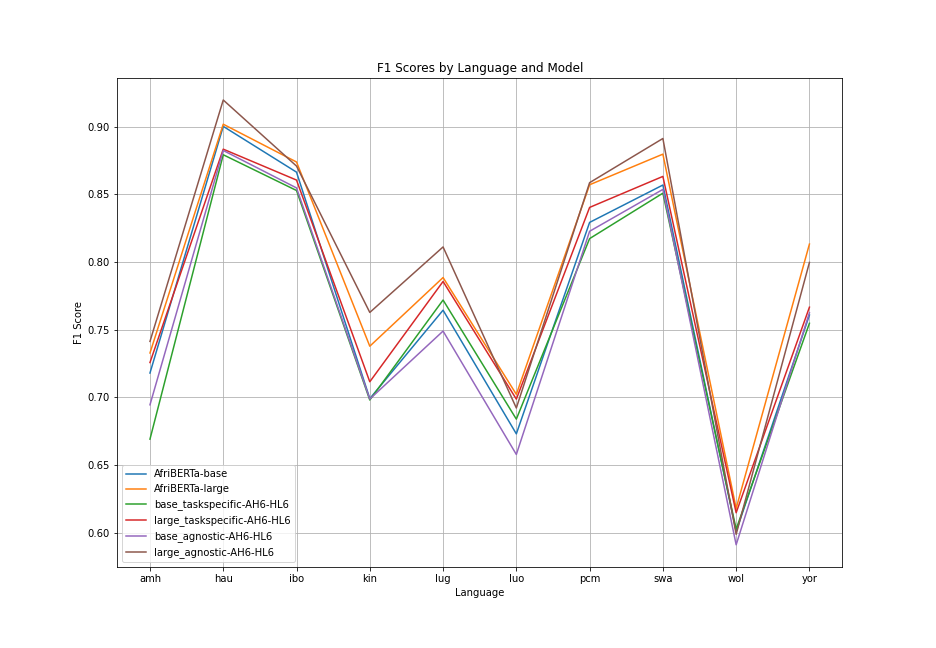}
    \label{fig:distilled_models}
\end{figure*}

\section{Pruning}
Unstructured magnitude pruning involves setting a binary mask, $M$, that determines which weights in the network are pruned based on their magnitude relative to a pruning threshold, $t$. Specifically, we define the mask $M$ as:

$M_{ij} = \begin{cases}
    1 & \text{if } |W_{ij}| \geq t \\
    0 & \text{otherwise}
\end{cases}$

where $W_{i,j}$ is the weight at position $(i,j)$ in the weight matrix, and $t$ is a threshold determined by the desired sparsity level.

\subsection{Average performances before and after fine-tuning}
\label{sec:appendixpruninga}

\begin{table*}
\centering
\caption{\textbf{Pruning before Finetuning:} Performance metrics of each language at all sparsity levels}
\label{tab:pruning_before_finetuning_table}
\begin{tabular}{|l|l|rrrr|rr|}
\hline
 & & \textbf{loss} & \textbf{precision} & \textbf{recall} & \textbf{f1} & \textbf{inference\_time} & \textbf{pruned\_params} \\
\textbf{prune\_rate} & \textbf{lang} & & & & & & \\
\hline
\textbf{0.10} & \textbf{amh} & 0.38 & 0.71 & 0.76 & 0.74 & 1.36 & 7078579 \\
& \textbf{hau} & 0.15 & 0.88 & 0.93 & 0.90 & 1.83 & 7078579 \\
& \textbf{ibo} & 0.20 & 0.86 & 0.88 & 0.87 & 1.85 & 7078579 \\
& \textbf{kin} & 0.35 & 0.72 & 0.78 & 0.75 & 1.73 & 7078579 \\
& \textbf{lug} & 0.29 & 0.78 & 0.82 & 0.80 & 1.24 & 7078579 \\
& \textbf{luo} & 0.44 & 0.70 & 0.73 & 0.71 & 0.57 & 7078579 \\
& \textbf{pcm} & 0.15 & 0.84 & 0.87 & 0.86 & 1.76 & 7078579 \\
& \textbf{swa} & 0.20 & 0.86 & 0.90 & 0.88 & 1.76 & 7078579 \\
& \textbf{wol} & 0.37 & 0.66 & 0.60 & 0.63 & 1.52 & 7078579 \\
& \textbf{yor} & 0.24 & 0.78 & 0.83 & 0.81 & 2.01 & 7078579 \\
\hline
\textbf{0.20} & \textbf{amh} & 0.37 & 0.71 & 0.76 & 0.73 & 1.25 & 14157158 \\
& \textbf{hau} & 0.16 & 0.87 & 0.93 & 0.90 & 1.56 & 14157158 \\
& \textbf{ibo} & 0.20 & 0.85 & 0.89 & 0.87 & 1.80 & 14157158 \\
& \textbf{kin} & 0.34 & 0.70 & 0.78 & 0.74 & 1.73 & 14157158 \\
& \textbf{lug} & 0.29 & 0.78 & 0.82 & 0.80 & 1.12 & 14157158 \\
& \textbf{luo} & 0.46 & 0.67 & 0.72 & 0.69 & 0.58 & 14157158 \\
& \textbf{pcm} & 0.16 & 0.83 & 0.87 & 0.85 & 1.62 & 14157158 \\
& \textbf{swa} & 0.21 & 0.86 & 0.90 & 0.88 & 1.61 & 14157158 \\
& \textbf{wol} & 0.37 & 0.63 & 0.60 & 0.61 & 1.41 & 14157158 \\
& \textbf{yor} & 0.26 & 0.78 & 0.83 & 0.81 & 1.79 & 14157158 \\
\hline
\textbf{0.30} & \textbf{amh} & 0.37 & 0.69 & 0.76 & 0.72 & 1.24 & 21235738 \\
& \textbf{hau} & 0.16 & 0.88 & 0.93 & 0.90 & 1.57 & 21235738 \\
& \textbf{ibo} & 0.19 & 0.86 & 0.89 & 0.87 & 1.68 & 21235738 \\
& \textbf{kin} & 0.35 & 0.70 & 0.78 & 0.74 & 1.62 & 21235738 \\
& \textbf{lug} & 0.29 & 0.77 & 0.82 & 0.79 & 1.14 & 21235738 \\
& \textbf{luo} & 0.46 & 0.69 & 0.72 & 0.70 & 0.60 & 21235738 \\
& \textbf{pcm} & 0.15 & 0.83 & 0.86 & 0.85 & 1.62 & 21235738 \\
& \textbf{swa} & 0.21 & 0.86 & 0.90 & 0.88 & 1.80 & 21235738 \\
& \textbf{wol} & 0.36 & 0.64 & 0.60 & 0.62 & 1.38 & 21235738 \\
& \textbf{yor} & 0.25 & 0.79 & 0.83 & 0.81 & 1.80 & 21235738 \\
\hline
\textbf{0.40} & \textbf{amh} & 0.38 & 0.70 & 0.76 & 0.73 & 1.27 & 28314317 \\
& \textbf{hau} & 0.16 & 0.86 & 0.93 & 0.90 & 1.69 & 28314317 \\
& \textbf{ibo} & 0.19 & 0.86 & 0.89 & 0.87 & 1.66 & 28314317 \\
& \textbf{kin} & 0.35 & 0.70 & 0.78 & 0.73 & 1.63 & 28314317 \\
& \textbf{lug} & 0.28 & 0.77 & 0.82 & 0.79 & 1.12 & 28314317 \\
& \textbf{luo} & 0.44 & 0.68 & 0.71 & 0.70 & 0.53 & 28314317 \\
& \textbf{pcm} & 0.15 & 0.84 & 0.87 & 0.85 & 1.73 & 28314317 \\
& \textbf{swa} & 0.21 & 0.85 & 0.89 & 0.87 & 1.61 & 28314317 \\
& \textbf{wol} & 0.37 & 0.65 & 0.59 & 0.62 & 1.38 & 28314317\\
& \textbf{yor} & 0.26 & 0.78 & 0.82 & 0.80 & 1.80 & 28314317\\
\hline
\textbf{0.50} & \textbf{amh} & 0.37 & 0.69 & 0.76 & 0.72 & 1.36 & 35392896\\
& \textbf{hau} & 0.17 & 0.87 & 0.93 & 0.90 & 1.56 & 35392896\\
& \textbf{ibo} & 0.21 & 0.85 & 0.88 & 0.87 & 1.67 & 35392896\\
& \textbf{kin} & 0.34 & 0.70 & 0.78 & 0.74 & 1.65 & 35392896\\
& \textbf{lug} & 0.29 & 0.76 & 0.81 & 0.78 & 1.30 & 35392896\\
& \textbf{luo} & 0.44 & 0.68 & 0.71 & 0.70 & 0.52 & 35392896\\
& \textbf{pcm} & 0.16 & 0.82 & 0.86 & 0.84 & 1.74 & 35392896\\
& \textbf{swa} & 0.22 & 0.84 & 0.89 & 0.86 & 1.63 & 35392896\\
& \textbf{wol} & 0.37 & 0.64 & 0.59 & 0.61 & 1.38 & 35392896\\
& \textbf{yor} & 0.26 & 0.78 & 0.82 & 0.80 & 1.79 & 35392896\\
\hline
\end{tabular}
\end{table*}%

\begin{table*}
\centering
\begin{tabular}{|l|l|rrrr|rr|}
\hline
 \textbf{prune\_rate} & \textbf{lang} & \textbf{loss} & \textbf{precision} & \textbf{recall} & \textbf{f1} & \textbf{inference\_time} & \textbf{pruned\_params} \\
\hline
\textbf{0.60} & \textbf{amh} & 0.39 & 0.68 & 0.75 & 0.71 & 1.25 & 42471475\\
& \textbf{hau} & 0.18 & 0.86 & 0.92 & 0.89 & 1.57 & 42471475\\
& \textbf{ibo} & 0.19 & 0.85 & 0.88 & 0.87 & 1.63 & 42471475\\
& \textbf{kin} & 0.36 & 0.69 & 0.78 & 0.73 & 1.60 & 42471475\\
& \textbf{lug} & 0.29 & 0.76 & 0.81 & 0.78 & 1.20 & 42471475\\
& \textbf{luo} & 0.43 & 0.68 & 0.70 & 0.69 & 0.58 & 42471475\\
& \textbf{pcm} & 0.16 & 0.82 & 0.85 & 0.84 & 1.75 & 42471475\\
& \textbf{swa} & 0.21 & 0.84 & 0.89 & 0.86 & 1.67 & 42471475\\
& \textbf{wol} & 0.37 & 0.62 & 0.59 & 0.61 & 1.41 & 42471475\\
& \textbf{yor} & 0.28 & 0.78 & 0.81 & 0.80 & 1.78 & 42471475\\
\hline
\textbf{0.70} & \textbf{amh} & 0.40 & 0.66 & 0.73 & 0.69 & 1.30 & 49550054\\
& \textbf{hau} & 0.19 & 0.85 & 0.92 & 0.88 & 1.55 & 49550054\\
& \textbf{ibo} & 0.20 & 0.83 & 0.87 & 0.85 & 1.69 & 49550054\\
& \textbf{kin} & 0.37 & 0.68 & 0.76 & 0.71 & 1.59 & 49550054\\
& \textbf{lug} & 0.29 & 0.73 & 0.80 & 0.76 & 1.14 & 49550054\\
& \textbf{luo} & 0.44 & 0.65 & 0.69 & 0.67 & 0.52 & 49550054\\
& \textbf{pcm} & 0.17 & 0.80 & 0.85 & 0.83 & 1.76 & 49550054  \\ 
& \textbf{swa} & 0.21 & 0.83 & 0.88 & 0.86 & 1.77 & 49550054  \\
& \textbf{wol} & 0.38 & 0.57 & 0.56 & 0.56 & 1.43 & 49550054 \\
& \textbf{yor} & 0.29 & 0.76 & 0.79 & 0.77 & 1.85 & 49550054 \\
\hline
\textbf{0.80} & \textbf{amh} & 0.43 & 0.62 & 0.70 & 0.66 & 1.25 & 56628634 \\
& \textbf{hau} & 0.20 & 0.83 & 0.90 & 0.87 & 1.55 & 56628634 \\
& \textbf{ibo} & 0.21 & 0.81 & 0.84 & 0.83 & 1.67 & 56628634 \\
& \textbf{kin} & 0.41 & 0.62 & 0.71 & 0.66 & 1.63 & 56628634 \\
& \textbf{lug} & 0.32 & 0.69 & 0.76 & 0.72 & 1.14 & 56628634 \\
& \textbf{luo} & 0.47 & 0.58 & 0.63 & 0.60 & 0.59 & 56628634 \\
& \textbf{pcm} & 0.19 & 0.76 & 0.82 & 0.79 & 1.60 & 56628634 \\
& \textbf{swa} & 0.22 & 0.80 & 0.87 & 0.83 & 1.66 & 56628634 \\
& \textbf{wol} & 0.39 & 0.53 & 0.53 & 0.53 & 1.38 & 56628634 \\
& \textbf{yor} & 0.32 & 0.72 & 0.76 & 0.74 & 1.78 & 56628634 \\
\hline
\textbf{0.90} & \textbf{amh} & 0.50 & 0.55 & 0.64 & 0.59 & 1.24 & 63707213 \\
& \textbf{hau} & 0.22 & 0.80 & 0.88 & 0.84 & 1.61 & 63707213 \\
& \textbf{ibo} & 0.24 & 0.79 & 0.83 & 0.81 & 1.71 & 63707213 \\
& \textbf{kin} & 0.45 & 0.57 & 0.68 & 0.62 & 1.61 & 63707213 \\
& \textbf{lug} & 0.36 & 0.63 & 0.72 & 0.67 & 1.20 & 63707213 \\
& \textbf{luo} & 0.53 & 0.50 & 0.58 & 0.53 & 0.52 & 63707213 \\
& \textbf{pcm} & 0.24 & 0.69 & 0.77 & 0.73 & 1.76 & 63707213 \\
& \textbf{swa} & 0.24 & 0.77 & 0.85 & 0.81 & 1.70 & 63707213 \\
& \textbf{wol} & 0.44 & 0.43 & 0.46 & 0.45 & 1.50 & 63707213 \\
& \textbf{yor} & 0.39 & 0.62 & 0.70 & 0.66 & 1.96 & 63707213 \\
\hline
\textbf{0.95} & \textbf{amh} & 0.51 & 0.50 & 0.61 & 0.55 & 1.24 & 67246502 \\
& \textbf{hau} & 0.23 & 0.77 & 0.86 & 0.81 & 1.56 & 67246502 \\
& \textbf{ibo} & 0.25 & 0.75 & 0.81 & 0.78 & 1.66 & 67246502 \\
& \textbf{kin} & 0.44 & 0.53 & 0.65 & 0.59 & 1.62 & 67246502 \\
& \textbf{lug} & 0.37 & 0.58 & 0.69 & 0.63 & 1.15 & 67246502 \\
& \textbf{luo} & 0.54 & 0.42 & 0.52 & 0.47 & 0.53 & 67246502 \\
& \textbf{pcm} & 0.27 & 0.64 & 0.74 & 0.69 & 1.66 & 67246502 \\
& \textbf{swa} & 0.90 & 0.50 & 0.59 & 0.53 & 1.65 & 67246502 \\
& \textbf{wol} & 1.05 & 0.28 & 0.32 & 0.29 & 1.42 & 67246502 \\
& \textbf{yor} & 1.02 & 0.37 & 0.46 & 0.40 & 1.82 & 67246502 \\
\hline
\end{tabular}
\end{table*}

\begin{table*}
\centering
\caption{\textbf{Pruning after Finetuning:} Performance metrics of each language at all sparsity levels}
\label{tab:pruning_after_finetuning_table}
\begin{tabular}{|l|l|rrrr|rr|}
\hline
 & & \textbf{loss} & \textbf{precision} & \textbf{recall} & \textbf{f1} & \textbf{inference\_time} & \textbf{pruned\_params} \\
\textbf{prune\_rate} & \textbf{lang} & & & & & & \\
\hline
\textbf{0.10} & \textbf{amh} & 0.37 & 0.72 & 0.76 & 0.74 & 1.88 & 7078579 \\
& \textbf{hau} & 0.17 & 0.87 & 0.93 & 0.90 & 1.93 & 7078579 \\
& \textbf{ibo} & 0.19 & 0.85 & 0.88 & 0.87 & 2.96 & 7078579 \\
& \textbf{kin} & 0.34 & 0.70 & 0.79 & 0.74 & 4.01 & 7078579 \\
& \textbf{lug} & 0.28 & 0.78 & 0.82 & 0.80 & 1.18 & 7078579 \\
& \textbf{luo} & 0.45 & 0.69 & 0.71 & 0.70 & 0.69 & 7078579 \\
& \textbf{pcm} & 0.15 & 0.84 & 0.87 & 0.85 & 1.76 & 7078579 \\
& \textbf{swa} & 0.20 & 0.86 & 0.90 & 0.88 & 1.76 & 7078579 \\
& \textbf{wol} & 0.36 & 0.63 & 0.60 & 0.61 & 1.64 & 7078579 \\
& \textbf{yor} & 0.25 & 0.78 & 0.83 & 0.80 & 2.35 & 7078579 \\
\hline
\textbf{0.20} & \textbf{amh} & 0.35 & 0.71 & 0.76 & 0.74 & 1.21 & 14157158 \\
 & \textbf{hau} & 0.16 & 0.87 & 0.93 & 0.90 & 1.65 & 14157158 \\
 & \textbf{ibo} & 0.19 & 0.85 & 0.88 & 0.87 & 1.75 & 14157158 \\
 & \textbf{kin} & 0.32 & 0.70 & 0.79 & 0.74 & 1.71 & 14157158 \\
 & \textbf{lug} & 0.27 & 0.77 & 0.82 & 0.80 & 1.18 & 14157158 \\
 & \textbf{luo} & 0.43 & 0.68 & 0.71 & 0.69 & 0.55 & 14157158 \\
 & \textbf{pcm} & 0.14 & 0.83 & 0.86 & 0.85 & 1.73 & 14157158 \\
 & \textbf{swa} & 0.19 & 0.86 & 0.90 & 0.88 & 1.60 & 14157158 \\
 & \textbf{wol} & 0.34 & 0.63 & 0.60 & 0.61 & 1.45 & 14157158 \\
 & \textbf{yor} & 0.24 & 0.78 & 0.83 & 0.80 & 1.94 & 14157158 \\
 \hline
\textbf{0.30} & \textbf{amh} & 0.32 & 0.72 & 0.76 & 0.73 & 1.21 & 21235738 \\
& \textbf{hau} & 0.14 & 0.87 & 0.93 & 0.90 & 1.66 & 21235738 \\
& \textbf{ibo} & 0.17 & 0.85 & 0.88 & 0.87 & 1.90 & 21235738 \\
& \textbf{kin} & 0.30 & 0.70 & 0.79 & 0.74 & 1.58 & 21235738 \\
& \textbf{lug} & 0.25 & 0.77 & 0.82 & 0.79 & 1.15 & 21235738 \\
& \textbf{luo} & 0.40 & 0.68 & 0.70 & 0.69 & 0.56 & 21235738 \\
& \textbf{pcm} & 0.13 & 0.82 & 0.86 & 0.84 & 1.58 & 21235738 \\
& \textbf{swa} & 0.18 & 0.85 & 0.90 & 0.88 & 1.73 & 21235738 \\
& \textbf{wol} & 0.32 & 0.63 & 0.59 & 0.61 & 1.51 & 21235738 \\
& \textbf{yor} & 0.22 & 0.78 & 0.83 & 0.81 & 2.04 & 21235738 \\
\hline
\textbf{0.40} & \textbf{amh} & 0.29 & 0.72 & 0.75 & 0.74 & 1.21 & 28314317 \\
& \textbf{hau} & 0.12 & 0.87 & 0.93 & 0.90 & 1.80 & 28314317 \\
& \textbf{ibo} & 0.15 & 0.85 & 0.88 & 0.87 & 1.62 & 28314317 \\
& \textbf{kin} & 0.26 & 0.69 & 0.78 & 0.74 & 1.57 & 28314317 \\
& \textbf{lug} & 0.22 & 0.77 & 0.81 & 0.79 & 1.18 & 28314317 \\
& \textbf{luo} & 0.36 & 0.69 & 0.69 & 0.69 & 0.51 & 28314317 \\
& \textbf{pcm} & 0.12 & 0.82 & 0.85 & 0.84 & 1.77 & 28314317 \\
& \textbf{swa} & 0.15 & 0.86 & 0.89 & 0.87 & 1.60 & 28314317 \\
& \textbf{wol} & 0.29 & 0.64 & 0.58 & 0.61 & 1.50 & 28314317 \\
& \textbf{yor} & 0.19 & 0.78 & 0.82 & 0.80 & 2.12 & 28314317 \\
\hline
\textbf{0.50} & \textbf{amh} & 0.23 & 0.71 & 0.74 & 0.72 & 1.32 & 35392896 \\
& \textbf{hau} & 0.10 & 0.86 & 0.93 & 0.89 & 1.51 & 35392896 \\
& \textbf{ibo} & 0.12 & 0.85 & 0.88 & 0.86 & 1.73 & 35392896 \\
& \textbf{kin} & 0.21 & 0.68 & 0.77 & 0.72 & 1.58 & 35392896 \\
& \textbf{lug} & 0.18 & 0.77 & 0.78 & 0.78 & 1.06 & 35392896 \\
& \textbf{luo} & 0.31 & 0.67 & 0.67 & 0.67 & 0.55 & 35392896 \\
& \textbf{pcm} & 0.11 & 0.81 & 0.83 & 0.82 & 1.68 & 35392896 \\
& \textbf{swa} & 0.12 & 0.85 & 0.89 & 0.87 & 1.66 & 35392896 \\
& \textbf{wol} & 0.25 & 0.64 & 0.55 & 0.59 & 1.49 & 35392896 \\
& \textbf{yor} & 0.17 & 0.79 & 0.80 & 0.79 & 1.95 & 35392896 \\
\hline
\end{tabular}
\end{table*}

\begin{table*}
\centering
\begin{tabular}{|l|l|rrrr|rr|}
\hline
 \textbf{prune\_rate} & \textbf{lang} & \textbf{loss} & \textbf{precision} & \textbf{recall} & \textbf{f1} & \textbf{inference\_time} & \textbf{pruned\_params} \\
\hline
\textbf{0.60} & \textbf{amh} & 0.19 & 0.71 & 0.66 & 0.68 & 1.32 & 42471475 \\
& \textbf{hau} & 0.09 & 0.83 & 0.88 & 0.86 & 1.70 & 42471475 \\
& \textbf{ibo} & 0.10 & 0.84 & 0.84 & 0.84 & 1.75 & 42471475 \\
& \textbf{kin} & 0.19 & 0.67 & 0.65 & 0.66 & 1.72 & 42471475 \\
& \textbf{lug} & 0.21 & 0.72 & 0.60 & 0.66 & 1.08 & 42471475 \\
& \textbf{luo} & 0.36 & 0.52 & 0.47 & 0.49 & 0.50 & 42471475 \\
& \textbf{pcm} & 0.11 & 0.77 & 0.76 & 0.76 & 1.60 & 42471475 \\
& \textbf{swa} & 0.10 & 0.84 & 0.85 & 0.84 & 1.63 & 42471475 \\
& \textbf{wol} & 0.23 & 0.64 & 0.42 & 0.50 & 1.49 & 42471475 \\
& \textbf{yor} & 0.18 & 0.76 & 0.66 & 0.71 & 1.94 & 42471475 \\
\hline
\textbf{0.70} & \textbf{amh} & 0.38 & 0.45 & 0.23 & 0.30 & 1.35 & 49550054 \\
& \textbf{hau} & 0.30 & 0.66 & 0.58 & 0.62 & 1.71 & 49550054 \\
& \textbf{ibo} & 0.31 & 0.66 & 0.50 & 0.57 & 1.73 & 49550054 \\
& \textbf{kin} & 0.38 & 0.63 & 0.22 & 0.32 & 1.60 & 49550054 \\
& \textbf{lug} & 0.47 & 0.48 & 0.11 & 0.18 & 1.13 & 49550054 \\
& \textbf{luo} & 0.63 & 0.23 & 0.09 & 0.12 & 0.58 & 49550054 \\
& \textbf{pcm} & 0.31 & 0.49 & 0.27 & 0.35 & 1.60 & 49550054 \\
& \textbf{swa} & 0.32 & 0.64 & 0.50 & 0.56 & 1.77 & 49550054 \\
& \textbf{wol} & 0.30 & 0.59 & 0.13 & 0.21 & 1.53 & 49550054 \\
& \textbf{yor} & 0.42 & 0.57 & 0.20 & 0.29 & 1.89 & 49550054 \\
\hline
\textbf{0.80} & \textbf{amh} & 1.25 & 0.23 & 0.04 & 0.06 & 1.27 & 56628634 \\
& \textbf{hau} & 1.39 & 0.48 & 0.33 & 0.35 & 1.54 & 56628634 \\
& \textbf{ibo} & 1.29 & 0.64 & 0.18 & 0.28 & 1.78 & 56628634 \\
& \textbf{kin} & 1.31 & 0.59 & 0.13 & 0.21 & 1.71 & 56628634 \\
& \textbf{lug} & 1.25 & 0.35 & 0.02 & 0.04 & 1.20 & 56628634 \\
& \textbf{luo} & 1.57 & 0.15 & 0.05 & 0.05 & 0.57 & 56628634 \\
& \textbf{pcm} & 1.22 & 0.37 & 0.05 & 0.09 & 1.75 & 56628634 \\
& \textbf{swa} & 1.35 & 0.54 & 0.25 & 0.31 & 1.72 & 56628634 \\
& \textbf{wol} & 1.13 & 0.05 & 0.01 & 0.01 & 1.40 & 56628634 \\
& \textbf{yor} & 1.09 & 0.59 & 0.04 & 0.06 & 1.77 & 56628634 \\
\hline
\textbf{0.90} & \textbf{amh} & 2.02 & 0.01 & 0.02 & 0.06 & 1.24 & 63707213 \\
& \textbf{hau} & 2.02 & 0.03 & 0.05 & 0.18 & 1.75 & 63707213 \\
& \textbf{ibo} & 2.01 & 0.03 & 0.04 & 0.12 & 1.81 & 63707213 \\
& \textbf{kin} & 2.04 & 0.02 & 0.02 & 0.08 & 1.76 & 63707213 \\
& \textbf{lug} & 2.05 & 0.01 & 0.02 & 0.08 & 1.23 & 63707213 \\
& \textbf{luo} & 2.07 & 0.01 & 0.01 & 0.04 & 0.52 & 63707213 \\
& \textbf{pcm} & 2.04 & 0.02 & 0.03 & 0.11 & 1.65 & 63707213 \\
& \textbf{swa} & 2.02 & 0.03 & 0.04 & 0.14 & 1.82 & 63707213 \\
& \textbf{wol} & 2.01 & 0.01 & 0.02 & 0.07 & 1.43 & 63707213 \\
& \textbf{yor} & 1.96 & 0.02 & 0.02 & 0.06 & 1.99 & 63707213 \\
\hline
\textbf{0.95} & \textbf{amh} & 2.19 & 0.01 & 0.09 & 0.02 & 1.37 & 67246502 \\
& \textbf{hau} & 2.19 & 0.01 & 0.12 & 0.02 & 1.62 & 67246502 \\
& \textbf{ibo} & 2.19 & 0.01 & 0.13 & 0.03 & 2.02 & 67246502 \\
& \textbf{kin} & 2.19 & 0.01 & 0.15 & 0.02 & 1.80 & 67246502 \\
& \textbf{lug} & 2.20 & 0.01 & 0.13 & 0.02 & 1.13 & 67246502 \\
& \textbf{luo} & 2.21 & 0.01 & 0.10 & 0.01 & 0.60 & 67246502 \\
& \textbf{pcm} & 2.20 & 0.01 & 0.10 & 0.01 & 1.69 & 67246502 \\
& \textbf{swa} & 2.18 & 0.01 & 0.13 & 0.02 & 1.88 & 67246502 \\
& \textbf{wol} & 2.19 & 0.00 & 0.08 & 0.01 & 1.61 & 67246502 \\
& \textbf{yor} & 2.19 & 0.01 & 0.10 & 0.01 & 1.90 & 67246502 \\
\hline
\end{tabular}
\end{table*}

\begin{figure*}[t!]
\centering
\subfigure[Pruning before fine-tuning]{
    \includegraphics[width=0.45\textwidth]{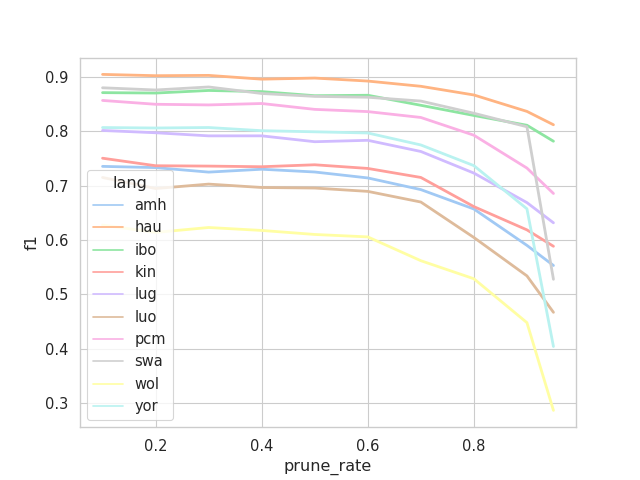}
}
\hfill
\subfigure[Pruning after fine-tuning]{
    \includegraphics[width=0.45\textwidth]{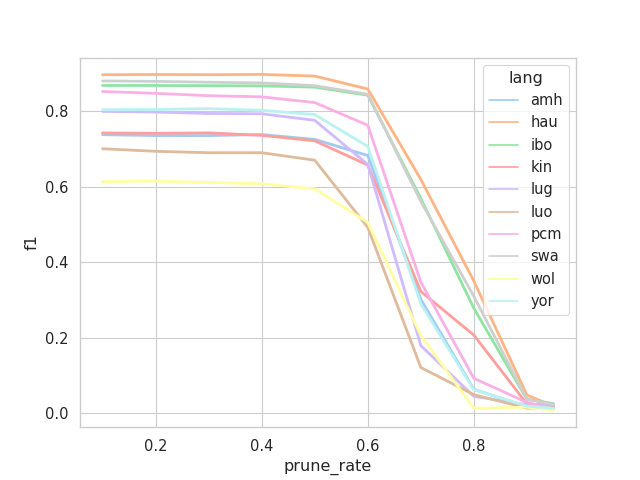}
}
\caption{Pruning \textbf{before vs after} fine-tuning: F1 scores averaged across the performances of each language.}
\label{fig:before_vs_after_finetuning_2}
\end{figure*}

\begin{figure*}[t!]
\centering
\subfigure[Inference Time wrt Sparsity Level]{
    \includegraphics[width=0.45\textwidth]{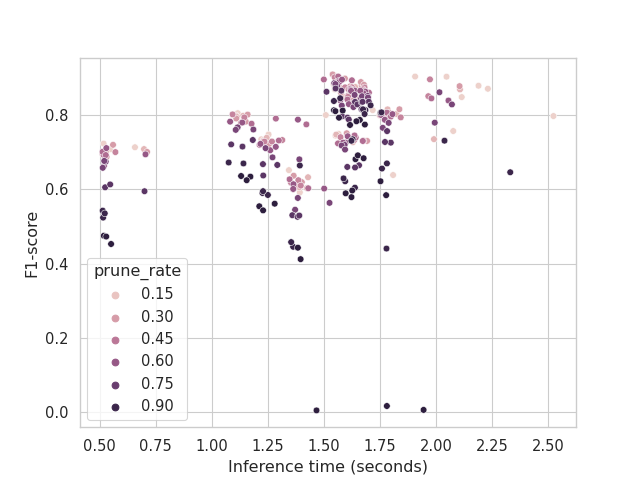}
}
\hfill
\subfigure[Inference Time wrt Language Groups]{
    \includegraphics[width=0.45\textwidth]{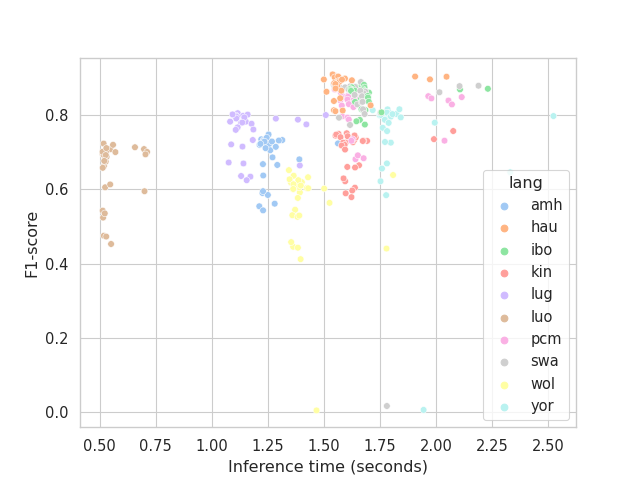}
}
\caption{Comparison of how Sparsity Level and Language Groups affect Inference Time.}
\label{fig:inference_time}
\end{figure*}

This section details the performance scores of all language tasks on all sparsity levels, before and after fine-tuning. See Tables \ref{tab:pruning_before_finetuning_table} \& \ref{tab:pruning_after_finetuning_table} and Figures \ref{fig:before_vs_after_finetuning_2} \& \ref{fig:inference_time} for references.

\subsection{The Impact of Pruning on Cross-Lingual Transfer}
\label{sec:appendixpruningb}

\begin{figure*}[t!]
\centering
\subfigure[Mean F1 scores of languages over sparsity levels]{
    \includegraphics[width=0.45\textwidth]{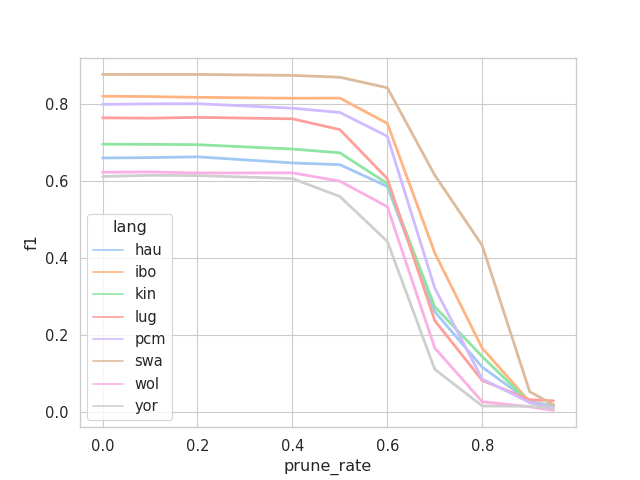}
}
\hfill
\subfigure[Distribution of the languages F1-scores over sparsity levels]{
    \includegraphics[width=0.45\textwidth]{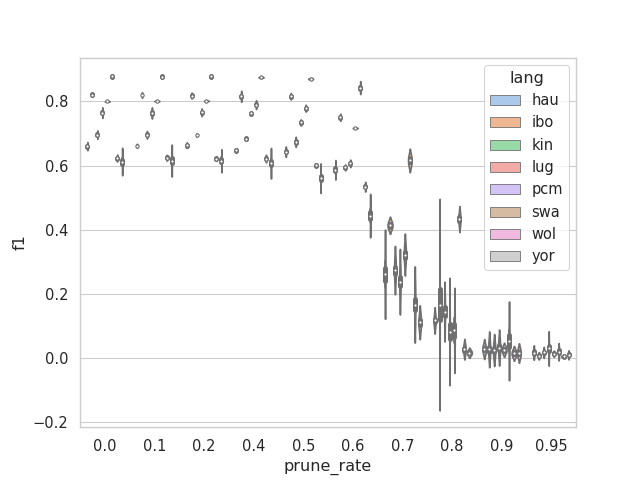}
}
\caption{\textbf{OOD Generalization:} Performances stay consistent with dense models up till 50\% sparsity.}
\label{fig:ood_performance}
\end{figure*}

This section analyses the impact of pruning on the cross-lingual transfer learning capabilities of the AfriBERTa model.

\paragraph{Few-shot learning} To evaluate the effectiveness of the AfriBERTa model in capturing linguistic intricacies in "unknown" languages, we fine-tuned it on two low-resource African languages - Fon and Bambara - from MasakhaNER 2.0. Results show that the performances when pruning after fine-tuning were comparable to the performances of the known languages. However, we observed that inference time increased by 2 to 3 times the usual inference time, and performances deteriorated rapidly from 70\% sparsity downwards. These observations suggest that the AfriBERTa model might have leveraged linguistic similarities and relationships inherent in languages sharing geographical regions.

\paragraph{Knowledge transfer from low-resource to high-resource} Our results show that AfriBERTa leverages the linguistic similarities and relationships inherent in languages that share geographical regions \citep{ogueji-etal-2021-small}. To test this assumption, we performed downstream on a high-resource language, Chinese, from the MSRA NER dataset \citep{feng2006chinese}. Surprisingly, AfriBERTa performed very strongly, implying that cross-lingual transfer is possible even between languages with no known affiliation. However, inference time is about four times (5–7 sec) the usual time it takes to perform inference on familiar language data.

\paragraph{Zero-shot learning} We further explored the zero-shot transferability of AfriBERTa on an unseen-before language at different levels of sparsity. Our findings reveal that F1 scores ranging between 40\% and 60\% remain competitive with MasakhaNER 2.0's zero-shot experiments on Afro-XLMR\citep{alabi_adapting_2022}, even at 50\% sparsity. Predicting on a new language from models trained on languages from the same geographical region seems to perform zero-shot more confidently, consistent with findings in MasakhaNER 2.0's experiments \citep{adelani2022masakhaner}. However, we observed a considerable drop in performance at 70\% sparsity, indicating that pruning might not be suitable for zero-shot learning in low-resource settings beyond a certain threshold.

\subsection{Impact of pruning on inference time}
\label{sec:appendixpruningc}
The effect of pruning on inference time is a significant component of our research, and we discovered that when the sparsity level grows, inference time decreases noticeably. However, our investigation in Figure \ref{fig:inference_time} indicated a wide range of inference time disparities, which might be related to variables other than the pruning rate itself. Our findings indicate that language-specific characteristics may have a considerable impact on inference time, as performance indicators showed no true connection with inference time. Further investigation is needed to study the language-specific aspects influencing inference time and to establish the ideal sparsity values for each language, taking both performance metrics and inference time into account.

\end{document}